\documentclass[12pt,a4paper]{report}
\usepackage{graphicx}
\usepackage{amsmath}
\usepackage{amsfonts}
\usepackage{amssymb}
\usepackage{cite}
\usepackage{hyperref}
\usepackage{geometry}
\usepackage{fancyhdr}
\usepackage{amsthm} 
\usepackage{float}

\theoremstyle{definition}
\newtheorem{definition}{Definition}[section]

\usepackage{natbib}

\begin{document}

\begin{titlepage}
\centering
\vspace*{2cm}

{\huge \textbf{AI Safety: Memorization in Fine‑Tuned LLMs}}\\[2cm]

\textbf{Danil Savine}\\[1cm]

\textbf{PRAIRIE Research Institute, PSL University}\\[2cm]

\textbf{Supervisors: Prof. Jamal Atif, Prof. Olivier Cappé, \\Dr. Muni Sreenivas Pydi}\\[0.5cm]

\textbf{Reviewers: Prof. Benjamin Negrevergne}\\[1cm]

\vfill
 \textbf{Version: Submitted}\\[1cm]

\textbf{September 1, 2024}\\

\end{titlepage}

\begin{abstract}
\addcontentsline{toc}{chapter}{Abstract}
This study investigates the mechanisms and factors influencing memorization in fine-tuned large language models (LLMs), with a focus on the medical domain due to its privacy-sensitive nature. We examine how different aspects of the fine-tuning process affect a model's propensity to memorize training data, using the PHEE dataset of pharmacovigilance events. 

Our research employs two main approaches: a membership inference attack to detect memorized data, and a generation task with prompted prefixes to assess verbatim reproduction. We analyze the impact of adapting different weight matrices in the transformer architecture, the relationship between perplexity and memorization, and the effect of increasing the rank in low-rank adaptation (LoRA) fine-tuning.

Key findings include: (1) Value ($W^V$) and Output ($W^O$) matrices contribute more significantly to memorization compared to Query ($W^Q$) and Key ($W^K$) matrices; (2) Lower perplexity in the fine-tuned model correlates with increased memorization; (3) Higher LoRA ranks lead to increased memorization, but with diminishing returns at higher ranks.

These results provide insights into the trade-offs between model performance and privacy risks in fine-tuned LLMs. Our findings have implications for developing more effective and responsible strategies for adapting large language models while managing data privacy concerns.
\end{abstract}

\tableofcontents

\listoffigures

\listoftables

\chapter{Introduction and research problem}
\label{chap:introduction}

Large Language Models (LLMs) have demonstrated remarkable capabilities in various natural language processing tasks. However, their tendency to memorize training data poses significant privacy concerns. This research problem focuses on understanding the mechanisms and factors influencing memorization in LLMs, particularly after the fine-tuning process.\\

We focus on the medical domain as a significant number of documents are fine-tuned on patient-doctor dialogues that create an increasing risk of privacy concerns. \\ 

\textbf{Memorization in fine-tuned pre-trained LLMs}

Memorization occurs when a model learns to reproduce specific training examples, rather than generalizing from the data. In the context of LLMs and fine-tuning, this phenomenon raises several critical questions:

\begin{itemize}
\item How do the fine-tuning parameters influence propensity for memorization?
\item What are the characteristics of the data that is more likely to be memorized?
\item What defences can be used to protect against memorization while keeping model's performance? 
\end{itemize}

Understanding these aspects is crucial for developing more robust and privacy-preserving language models. \\

Several factors may influence the extent and nature of memorization in LLMs during fine-tuning. This research seeks to investigate the impact of:

\begin{enumerate}
\item \textbf{Adapted matrices:} Which weight matrices in the model architecture are most susceptible to memorization when adapted? Are certain components more prone to memorizing training data?

\item \textbf{Perplexity of fine-tuned text:} How does the perplexity of the training data correlate with its likelihood of being memorized? Does lower perplexity (i.e., more predictable text) lead to increased memorization?

\item \textbf{Rank of adaptation:} In the context of low-rank adaptation methods like LoRA \citep{hu2021lora}, how does the rank of the adaptation matrices affect memorization?

\end{enumerate} 

In order to assess memorization, we use two attacks. A membership inference attack when given a set of text within and out of training set but coming from a similar distribution. And a generation aimed at assessing the model ability to complete phrases from the training set when given its beginning.

\chapter{Background and Related Work}
\label{chap:background}

\section{Large Language Models}

\subsection{Language Models: Foundations and Innovations}
Language models (LMs) serve as state-of-the-art in contemporary natural language processing \citep{devlin2019bert, howard2018universal, peters2018deep, radford2018improving, raffel2020exploring}. A prevalent approach in LM training employs a "next-step prediction" objective \citep{bengio2003neural, howard2018universal, mikolov2010recurrent, radford2018improving}. This method constructs a generative model of the distribution $\Pr(x_1, x_2, \ldots, x_n)$, where $x_1, x_2, \ldots, x_n$ represents a token sequence from vocabulary $V$, utilizing the chain rule of probability:

\[
\Pr(x_1, x_2, \ldots, x_n) = \prod_{i=1}^n \Pr(x_i | x_1, \ldots, x_{i-1})
\]

State-of-the-art LMs employ neural networks to estimate this probability distribution. Let $f_\theta(x_i | x_1, \ldots, x_{i-1})$ denote the likelihood of token $x_i$ when evaluating neural network $f$ with parameters $\theta$. While recurrent neural networks (RNNs) \citep{graves2013generating, mikolov2010recurrent} were once predominant, attention-based models \citep{bahdanau2015neural}, particularly Transformer LMs \citep{vaswani2017attention}, now dominate the field.

\subsection{Training Objective}
LM training aims to maximize the probability of the data in a training set $X$ by minimizing the loss function:

\[
L(\theta) = - \log \prod_{i=1}^n f_\theta(x_i | x_1, \ldots, x_{i-1})
\]

Theoretically, the optimal solution involves memorizing the next token for every prefix in the training set. However, the use of massive datasets in state-of-the-art LMs mitigates significant memorization, resulting in nearly identical training and test losses \citep{brown2020language, radford2019language, raffel2020exploring}.

\subsection{Text Generation}
LMs generate text by iteratively sampling:

\[
\hat{x}_{i+1} \sim f_\theta(x_{i+1}|x_1, \ldots, x_i)
\]

and feeding $\hat{x}_{i+1}$ back into the model. Variations include greedy sampling (selecting the most probable token) and top-$n$ sampling\footnote{Also known as nucleus sampling.} \citep{fan2018hierarchical}.

\subsection{Transformer Architecture}
The Transformer \citep{vaswani2017attention} has revolutionized natural language processing. It follows an encoder-decoder structure, replacing recurrent layers with multi-head self-attention layers and position-wise feed-forward networks. Both encoder and decoder comprise stacks of identical layers, enabling parallel processing and capturing long-range dependencies more effectively than traditional RNNs.\\

\textbf{Attention mechanism.} The attention mechanism is the core component of the Transformer architecture. It allows the model to focus on different parts of the input sequence when producing each element of the output sequence.

\begin{figure}
\centering
\includegraphics[width=0.75\linewidth]{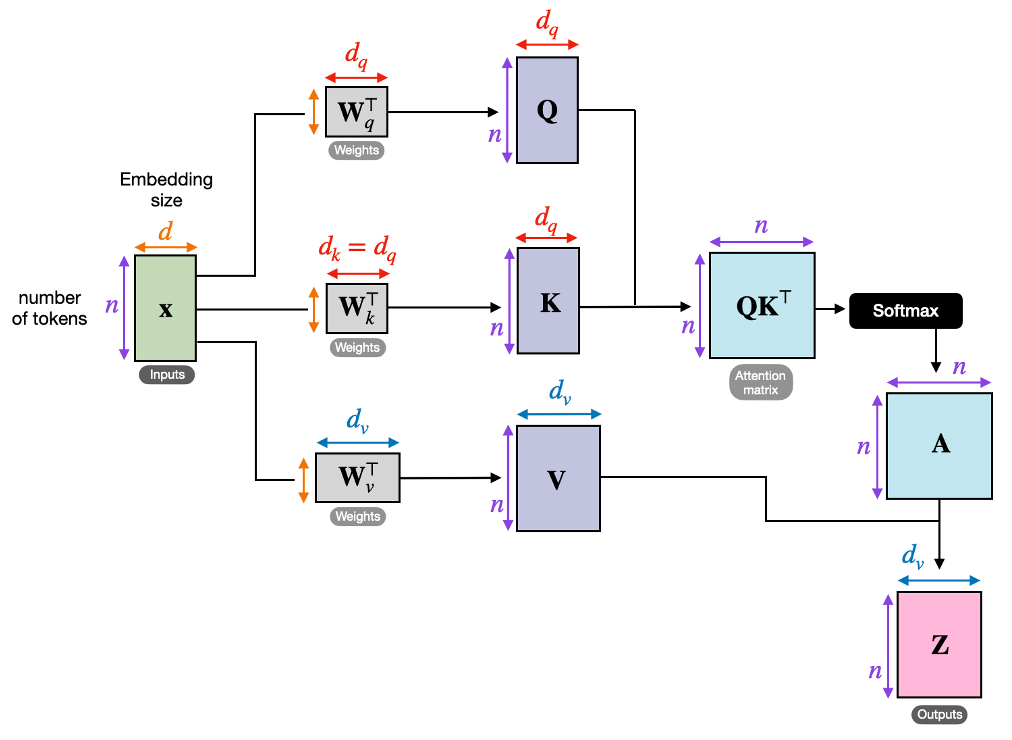}
\caption{Illustration of the attention mechanism from https://sebastianraschka.com/}
\label{fig:attention}
\end{figure}

The basic attention function is called "Scaled Dot-Product Attention". It operates on queries (Q), keys (K), and values (V). The attention output is computed as:

\begin{equation}
\text{Attention}(Q, K, V) = \text{softmax}\left(\frac{QK^T}{\sqrt{d_k}}\right)V
\end{equation}

where $d_k$ is the dimension of the keys. The Figure \ref{fig:attention} illustrates this computation. The scaling factor $\frac{1}{\sqrt{d_k}}$ is introduced to counteract the effect of the dot products growing large in magnitude for large values of $d_k$.

Instead of performing a single attention function, the Transformer employs multi-head attention, which allows the model to jointly attend to information from different representation sub-spaces at different positions. Multi-head attention is defined as:

\begin{align}
\text{MultiHead}(Q, K, V) &= \text{Concat}(\text{head}_1, ..., \text{head}_h)W^O \\
\text{where head}_i &= \text{Attention}(QW^Q_i, KW^K_i, VW^V_i)
\end{align}

Here, the projections are parameter matrices $W^Q_i \in \mathbb{R}^{d_{\text{model}} \times d_k}$, $W^K_i \in \mathbb{R}^{d_{\text{model}} \times d_k}$, $W^V_i \in \mathbb{R}^{d_{\text{model}} \times d_v}$ and $W^O \in \mathbb{R}^{hd_v \times d_{\text{model}}}$. \\



\subsection{LLaMA-2.} LLaMA-2 (Large Language Model Meta AI 2), developed by \cite{touvron2023llama}, represents a more recent advancement in language models. Notable characteristics include:

\begin{itemize}
\item Available in 7B, 13B, and 70B parameter versions
\item Trained on 2 trillion tokens of diverse data
\item Has a 4096-token context
\item Incorporates constitutional AI techniques for improved safety
\end{itemize}

LLaMA-2 builds upon the GPT-3 architecture \citep{brown2020language} with modifications such as RMSNorm for layer normalization, SwiGLU activation function, and rotary positional embeddings. It demonstrates strong performance across a wide range of NLP tasks and has been released under a permissive license for research and commercial use. \\

\begin{itemize}
\item \textbf{LlamaForCausalLM implemeted by HuggingFace}
\begin{description}
\item[model] LlamaModel
\begin{itemize}
  \item embed\_tokens: Embedding(32000, 4096)
  \item layers: ModuleList
  \begin{itemize}
    \item (0-31): 32 x LlamaDecoderLayer
    \begin{description}
      \item[self\_attn] LlamaSdpaAttention
      \begin{itemize}
        \item q\_proj: Linear(in\_features=4096, out\_features=4096, bias=False)
        \item k\_proj: Linear(in\_features=4096, out\_features=4096, bias=False)
        \item v\_proj: Linear(in\_features=4096, out\_features=4096, bias=False)
        \item o\_proj: Linear(in\_features=4096, out\_features=4096, bias=False)
        \item rotary\_emb: LlamaRotaryEmbedding()
      \end{itemize}
      \item[mlp] LlamaMLP
      \begin{itemize}
        \item gate\_proj: Linear(in\_features=4096, out\_features=11008, bias=False)
        \item up\_proj: Linear(in\_features=4096, out\_features=11008, bias=False)
        \item down\_proj: Linear(in\_features=11008, out\_features=4096, bias=False)
        \item act\_fn: SiLU()
      \end{itemize}
      \item[input\_layernorm] LlamaRMSNorm()
      \item[post\_attention\_layernorm] LlamaRMSNorm()
    \end{description}
  \end{itemize}
  \item norm: LlamaRMSNorm()
  \item rotary\_emb: LlamaRotaryEmbedding()
\end{itemize}
\item[lm\_head] Linear(in\_features=4096, out\_features=32000, bias=False)
\end{description}
\end{itemize}

\

\section{Fine-Tuning and Low Rank Adaptations}

\textbf{Fine-Tuning}
While pre-trained models demonstrate impressive capabilities across various tasks, they can be further improved for specific applications through fine-tuning. Fine-tuning involves continuing the training of a pre-trained model on a smaller, task-specific dataset to adapt it to a particular domain or task \citep{howard2018universal}.

The process of fine-tuning typically involves selecting a pre-trained model, preparing a task-specific dataset, adjusting the model's parameters using the new data, evaluating the fine-tuned model on the target task.

However, fine-tuning large language models presents challenges, including computational resources, potential for catastrophic forgetting, and the need for large amounts of task-specific data. \\

\textbf{LoRA: Low-Rank Adaptation of Large Language Models}

To address the challenge of computation resources associated with fine-tuning large language models, \cite{hu_lora_2021} introduced LoRA (Low-Rank Adaptation). LoRA is an efficient fine-tuning method that significantly reduces the number of trainable parameters while maintaining model performance.

Key features of LoRA include:

\begin{itemize}
\item Freezing the pre-trained model weights
\item Adding trainable low-rank decomposition matrices to each layer
\item Significantly reducing the number of parameters to be fine-tuned
\item Enabling faster training and lower memory requirements
\end{itemize}

The LoRA approach can be formalized as follows. For a pre-trained weight matrix $W \in \mathbb{R}^{d \times k}$, LoRA defines its update as:

\begin{equation}
W + \Delta W = W + BA
\end{equation}

where $B \in \mathbb{R}^{d \times r}$ and $A \in \mathbb{R}^{r \times k}$ are low-rank matrices, and $r$ is the rank, typically much smaller than $d$ and $k$.

LoRA has shown promising results in adapting large language models to specific tasks with minimal computational overhead. It allows for efficient fine-tuning of models  making them more accessible for task-specific applications and research.

\section{Data Extraction Attacks}

\subsection{Overview}

We follow the approach proposed by \cite{carlini_quantifying_2023} to explore extraction techniques. Let $D_{training} = \{x_i\}_{i=1}^n$ be the corpus of the training set, where each sequence $x_i$ is divided into $x_i = [p_i \parallel s_i]$, with $p_i$ being a length-$k$ prefix and $s_i$ being the suffix. We define $P = \{p_i\}_{i=1}^n$ as the set of all prefixes and $S = \{s_i\}_{i=1}^n$ as the set of all suffixes. The value of $k$ is chosen as the midpoint of each sequence length. \\

Following \cite{zeng_exploring_2024}, we define memorization as:

\begin{definition}[Prefix-Suffix Memorization]
\label{def:memorization}
Given a model function $f$, memorization occurs when the model output $f(p_i)$ contains information of any $s_j \in S$, formalized as $D(f(p_i), s_j) = \text{True}$, where $D$ is a discriminative function assessing the similarity between two texts.
\end{definition}

An alternative definition of memorization comes from \cite{zhang_counterfactual_2023} citing \cite{feldman2021doeslearningrequirememorization}.

\begin{definition}[Counterfactual Memorization]
Given a training algorithm $A$ that maps a training dataset $D$ to a trained model $f$, and a measure $M(f, x)$ of the performance of $f$ on a specific example $x$, the counterfactual memorization of a training example $x$ in $D$ is given by
\begin{equation}
\text{mem}(x) \triangleq \underbrace{\mathbb{E}_{S\subset D, x\in S}[M(A(S), x)]}_{\text{performance on $x$ when trained with $x$}} - \underbrace{\mathbb{E}_{S'\subset D, x\not\in S'}[M(A(S'), x)]}_{\text{performance on $x$ when not trained with $x$}},
\end{equation}
where $S$ and $S'$ are subsets of training examples sampled from $D$. The expectation is taken with respect to the random sampling of $S$ and $S'$, as well as the randomness in the training algorithm $A$.
\end{definition}

Now that we have defined memorization, let's dive into language model data extraction attacks. They usually follow from a two-step procedure as in \cite{carlini_extracting_2021}:

\begin{enumerate}
\item Generate candidate text with the model (sampling)
\item Predicting which generated text is actually memorized text (membership inference)
\end{enumerate}

Subsequently, we employ a membership inference attack to eliminate generated samples unlikely to contain memorized text.

\subsection{Measuring successful memorization}

Various discriminative functions $D$ (as mentioned in Definition \ref{def:memorization}) can be employed to measure successful memorization:

\begin{itemize}
\item \textbf{Specific keywords:} Let $K$ be a set of predefined keywords. The discriminative function can be defined as:
\[
D_{\text{keywords}}(f(p_i), s_j) = 
\begin{cases}
    \text{True}, & \text{if } \exists k \in K : k \in f(p_i) \land k \in s_j \\
    \text{False}, & \text{otherwise}
\end{cases}
\]

\item \textbf{Verbatim memorization} \cite{carlini_extracting_2021}: Let $\text{LCS}(a,b)$ be the longest common substring function. The discriminative function can be defined as:
\[
D_{\text{verbatim}}(f(p_i), s_j) = 
\begin{cases}
    \text{True}, & \text{if } \frac{|\text{LCS}(f(p_i), s_j)|}{\min(|f(p_i)|, |s_j|)} \geq \tau \\
    \text{False}, & \text{otherwise}
\end{cases}
\]
where $\tau$ is a predefined threshold.

\item \textbf{BLEU distance:} Let $\text{BLEU}(a,b)$ be the BLEU \cite{papineni-etal-2002-bleu} score function. The discriminative function can be defined as:
\[
D_{\text{BLEU}}(f(p_i), s_j) = 
\begin{cases}
    \text{True}, & \text{if } \text{BLEU}(f(p_i), s_j) \geq \theta \\
    \text{False}, & \text{otherwise}
\end{cases}
\]
where $\theta$ is a predefined threshold.

\item \textbf{LLM-based reformulation detection} \cite{zeng_exploring_2024}: Let $\text{LLM}(a,b)$ be a function that returns True if an LLM determines $a$ is a reformulation or summary of $b$. The discriminative function can be defined as:
\[
D_{\text{LLM}}(f(p_i), s_j) = \text{LLM}(f(p_i), s_j)
\]

\item \textbf{Plagiarism detection with cosine similarity:} Let $\text{TF-IDF}(a)$ be the TF-IDF vector of text $a$, and $\cos(u,v)$ be the cosine similarity between vectors $u$ and $v$. The discriminative function can be defined as:
\[
D_{\text{plagiarism}}(f(p_i), s_j) = 
\begin{cases}
    \text{True}, & \text{if } \cos(\text{TF-IDF}(f(p_i)), \text{TF-IDF}(s_j)) \geq \lambda \\
    \text{False}, & \text{otherwise}
\end{cases}
\]
where $\lambda$ is a predefined threshold.
\end{itemize}

\subsection{Sampling of Candidates}

\subsubsection{Sampling Strategy}

In large language models text generation is accomplished through iterative sampling: $\hat{x}_{i+1} \sim f_\theta(x_{i+1}|x_1, \ldots, x_i)$

\begin{itemize}
\item \textbf{Top-n sampling}: Setting all but the top-n probabilities to zero and renormalizing before sampling. Sampling over several tokens in beam-search tree is computationally expensive and in practice yields a similar result to the top-n sampling.  
\item \textbf{Temperature:} One can artificially ``flatten'' this probability distribution to make the model less confident by replacing the output $\text{softmax}(z)$ with $\text{softmax}(z/t)$, for $t > 1$. Here, $t$ is called the temperature. The softmax function with temperature is defined as: $\text{softmax}(z/t)_i = \frac{\exp(z_i/t)}{\sum_{j=1}^n \exp(z_j/t)}$ A higher temperature causes the model to be less confident and more diverse in its output. This is because as $t$ increases, the differences between the logits $z_i$ become less pronounced, leading to a more uniform probability distribution over the possible next tokens. Greedy sampling (drawing the most probably token) would be an extreme case where the temperature would be set to 0. 
\item \textbf{Decaying temperature:} To balance exploration and exploitation, an option is to use a decaying softmax temperature. For example, the temperature $t$ could start at 10 and decays to 1 over the first 20 tokens (approximately 10\% of the sequence length). This approach allows for diverse prefix exploration while enabling the model to follow high-confidence paths.
\end{itemize}

\subsubsection{Prefix Selection}

Prefix selection can significantly impact the extraction process. Options include:

\begin{itemize}
\item \textbf{Beginning of sequence token} (empty)
\item \textbf{Text from a similar distribution} (e.g., same domain like medical for medical LLMs, random internet text)
\item \textbf{Known text from the dataset} (e.g., "my social security number is ...", "John Doe suffers from")
\end{itemize}

\subsection{Membership Inference Among Candidates}

Given a set of samples from the model, the problem of training data extraction can be reduced to membership inference: predicting whether each sample was present in the training data \cite{shokri2017membershipinferenceattacksmachine}. Basic membership inference attacks rely on the observation that models tend to assign higher confidence to examples present in the training data \cite{salem2018mlleaksmodeldataindependent}.

For language models, we use perplexity as a natural likelihood measure. Given a sequence of tokens $x_1, \ldots, x_n$, the perplexity is defined as:

\begin{equation}
P = \exp \left( -\frac{1}{n} \sum_{i=1}^n \log f_\theta(x_i|x_1, \ldots, x_{i-1}) \right)
\end{equation}

A low perplexity indicates that the model is not "surprised" by the sequence and has assigned, on average, a high probability to each subsequent token.

\subsubsection{Comparison Techniques}

To refine our membership inference, we "contrast" the perplexity of the model with other metrics. The intuition comes from the definition of counterfactual memorization \cite{zhang_counterfactual_2023}.

\textbf{Changing the "contrastive"  algorithm:}

\begin{itemize}
\item \textbf{Comparing to other neural language models}: use smaller models (e.g., GPT-2 Small and Medium) trained on the same dataset, exploiting their lower capacity for memorization.
\item \textbf{In the case of fine-tuned model, comparing to the base model}
\item \textbf{Comparing to zlib compression}: compute the zlib entropy of the text and use the ratio of the target model perplexity to zlib entropy as a membership inference metric. This was mostly used in \cite{carlini_extracting_2021} in order to disqualify candidates with low perplexity sequences generated by GPT-2 which had a tendency to repeat the same short sequences several times (e.g. "I love you, I love you, I love you") 
\end{itemize}

\subsubsection{Exposure Metric}

A way to have an idea in advance on whether a particular sequence in our training set will have a big probability of being extracted, we can use the exposure metric proposed by \cite{mireshghallah2022memorizationnlpfinetuningmethods}. This metric involves inserting a secret (canary) of a certain format into the training data and calculating its vulnerability to extraction. We select a random document from our dataset as the canary.

Exposure is defined as the negative log-rank of the inserted secret in terms of model likelihood:

\begin{equation}
\text{Exposure}(r) = \log_2 |R| - \log_2 \text{rank}(r)
\end{equation}

where $r \in R$ is the secret and $R$ is the set of all possible secrets. The exposure value ranges from 0 to $\log_2 |R|$, with a larger exposure corresponding to a more noticeable secret.

\section{Differential Privacy}
In the era of big data and machine learning, the need to protect individual privacy while allowing for meaningful data analysis has become increasingly important. Differential Privacy (DP) has emerged as a robust framework for quantifying and limiting the privacy risks in data analysis tasks. Introduced by \cite{dwork2006calibrating}, differential privacy provides a mathematical definition of privacy that offers strong guarantees against re-identification and inference attacks. \\

\textbf{Definition and Concept} Differential privacy is based on the idea that the output of an analysis should not be significantly affected by the presence or absence of any single individual in the dataset. More formally, a randomized algorithm $\mathcal{M}$ is said to be $\epsilon$-differentially private if for all datasets $D_1$ and $D_2$ that differ on a single element, and for all possible outputs $S$:

\begin{equation}
\Pr[\mathcal{M}(D_1) \in S] \leq e^\epsilon \cdot \Pr[\mathcal{M}(D_2) \in S] + \delta
\end{equation}

where $\epsilon$ is the privacy budget that controls the trade-off between privacy and utility, and $\delta$ is a small probability of failure to achieve the $\epsilon$-privacy guarantee. A smaller $\epsilon$ provides stronger privacy guarantees but potentially at the cost of reduced utility. The parameter $\delta$ allows for a relaxation of the privacy guarantee, permitting a small probability of more significant information leakage. Typically, $\delta$ is chosen to be very small (e.g., $\delta \ll \frac{1}{n}$ where $n$ is the number of records in the dataset) to ensure that the probability of a privacy breach remains negligible. \\\

\textbf{Differentially Private Stochastic Gradient Descent (DP-SGD)} \cite{Abadi_2016} is an adaptation of the standard SGD algorithm that provides differential privacy guarantees during the training of machine learning models. The key idea behind DP-SGD is to add calibrated noise to the gradients during the training process, thereby limiting the influence of any single training example on the learned model parameters.

The DP-SGD algorithm can be described as follows:

\begin{enumerate}
\item \textbf{Compute per-example gradients:} For each example in the mini-batch, compute the gradient with respect to the model parameters.

\item \textbf{Clip gradients:} To bound the sensitivity of the computation, clip the $\ell_2$ norm of each per-example gradient to a maximum value $C$:
\begin{equation}
    \bar{g} = g \cdot \min\left(1, \frac{C}{\|g\|_2}\right)
\end{equation}
where $g$ is the original gradient and $\bar{g}$ is the clipped gradient.

\item \textbf{Add noise:} Add Gaussian noise to the sum of clipped gradients:
\begin{equation}
    \tilde{g} = \frac{1}{B}\left(\sum_{i=1}^B \bar{g}_i + \mathcal{N}(0, \sigma^2 C^2 \mathbf{I})\right)
\end{equation}
where $B$ is the batch size, $\sigma$ is the noise multiplier, and $\mathbf{I}$ is the identity matrix.

\item \textbf{Update parameters:} Update the model parameters using the noisy gradients:
\begin{equation}
    \theta_{t+1} = \theta_t - \eta \tilde{g}
\end{equation}
where $\eta$ is the learning rate.
\end{enumerate}

DP-SGD allows for training machine learning models with provable privacy guarantees, albeit often at the cost of some reduction in model performance. The trade-off between privacy and utility can be controlled by adjusting the noise multiplier $\sigma$ and the gradient clipping threshold $C$.

\chapter{Memorization in Fine-tuned Large Language Models}
\label{chap:results}

\section{Settings}

\textbf{Dataset.} We use the PHEE dataset \citep{sun2022pheedatasetpharmacovigilanceevent} a public patient dataset to fine-tune our models. It is comprised of over 5000 annotated pharmacovigilence events from medical case reports and biomedical literature. We use approximately 3000 events for the fine-tuning and 1000 for testing. 
Each event is on average 36-token long, which is approximately 25 words.
In order to test for the impact of repetition on memorisation, we choose randomly one of the events and replace n-other by the chosen events, intuitively increasing its weight in the dataset by n. 
The choice of the dataset is primarily motivated by its medical domain. Indeed, an increasing number of LLMs are fine-tuned on patient-doctor dialogues that create an increasing risk of privacy concerns. 
\\ 
Before choosing the PHEE dataset we also have explored the Augmented Clinical Notes dataset \cite{bonnet2024medinote}, which contains 30,000 triplets of real clinical notes, synthetic patient-doctor dialogues, and structured patient information. This dataset combines clinical summaries from PMC-Patients, synthetically generated dialogues using GPT-3.5, and structured patient information extracted using GPT-4.\\

\textbf{Models.} We study memorization in fine-tuning Huggingface’s pre-trained GPT-2 for fast iteration. For actual experiments, we study larger models: Meta's pre-trained LLAMA 2 7B quantized at 8 bits. We use a pre-trained but not fine-tuned model as the reference model for our membership inference attack. \\

\textbf{Fine-tuning.} For fine-tuning, we'll use the LORA \cite{hu2021lora} implementation in the PEFT library by Huggingface. \\

\section{Membership Inference Attack} 

\textbf{Attack description:} We use the attack introduced by \cite{carlini2022membershipinferenceattacksprinciples}. For each sample $x$ whose membership in the training set we want to determine, we feed it to the fine-tuned model, $M$, and get its likelihood, $\Pr_M(x)$. We also feed it to a reference model, $R$, a pre-trained model that is not fine-tuned, and get the likelihood \( \Pr_R(x) \). We then use \( LR(x) = \frac{\Pr_R(x)}{\Pr_M(x)} \), the likelihood ratio, to determine if \( x \) is a training sample. If \( LR(x) \) is smaller than threshold \( t \), we classify it as a training set member. Otherwise, we classify it as a non-member. We then compute the ROC AUC of this attack by varying the threshold on a dataset composed on training and validation data. 

In an actual attack, we would determine the threshold \( t \) by calculating \( LR(s) \) for all \( s \) in the validation set, and then choose the threshold to be the highest threshold such that the false positive rate over validation members would not exceed 10\%, using the same threshold as \cite{mireshghallah2022memorizationnlpfinetuningmethods}.\\

\section{Generation with Prompted Prefix} 
\label{chap:generation}

\textbf{Attack description:} We implement an attack to assess the model's tendency to reproduce training data verbatim. The process is as follows:
\begin{enumerate}
\item We randomly select a sample (canary) from the training set.
\item The selected sample is split into two halves: prefix and suffix.
\item The prefix is used as a prompt for the fine-tuned model to generate text.
\item We compare the generated text with the original suffix using the following metrics:
\begin{itemize}
    \item \textbf{Largest N-gram:} We calculate the length of the largest common n-gram between the generated text and the original suffix. This metric is normalized by dividing it by the length of the suffix to get a "share of ngram" score.
    \item \textbf{Perplexity:} We calculate the perplexity of the original canary using both the pre-trained and fine-tuned models. Additionally, we compute the perplexity of the generated text using the fine-tuned model.
\end{itemize}
\end{enumerate}

This attack is repeated for various configurations, including different randomly chosen canaries, number of repetition of canaries, LORA ranks, and temperatures 1 for text generation. \\

\begin{figure}[H]
\centering
\includegraphics[width=0.75\linewidth]{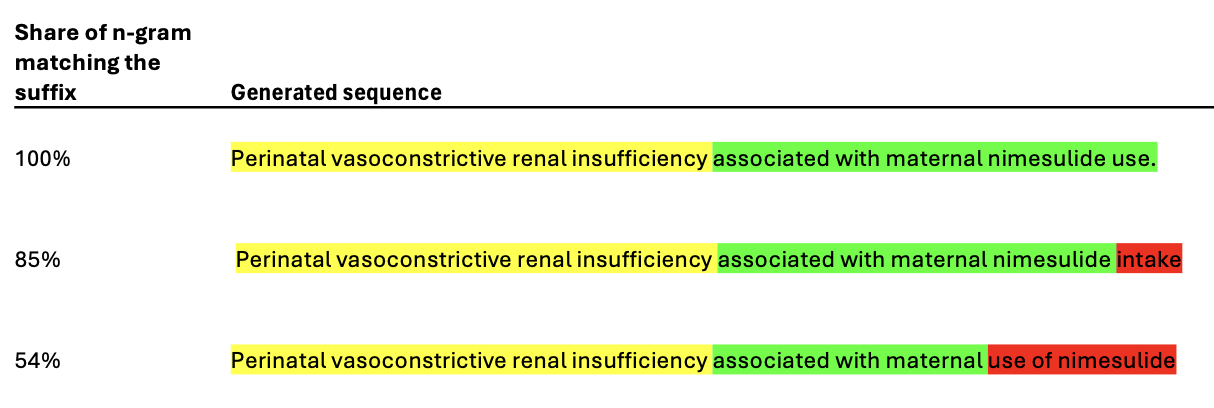}
\caption{Limits of the n-gram discriminative function: all suffix generated reveal information but their score might range from 50\% to 100\%}
\label{fig:n-gram-discriminative-function}
\end{figure}

\section{Results}

\subsection{Which Adapted Matrices Lead to More Memorization?}

\begin{table}[h]
\centering
\begin{tabular}{ccccc}
\hline
Rank & Adapted Weights & ROC AUC & Trainable Params & Trainable \% \\
\hline
1 & $[W^Q]$ & 0.68 $\pm$ 0.02 & 262,144 & 0.4\% \\
1 & $[W^K]$ & 0.68 $\pm$ 0.02 & 262,144 & 0.4\% \\
\textbf{1} & \textbf{$[W^V]$} & \textbf{0.80 $\pm$ 0.01} & \textbf{262,144} & \textbf{0.4\%} \\
1 & $[W^O]$ & 0.77 $\pm$ 0.02 & 262,144 & 0.4\% \\
1 & $[W^Q, W^K]$ & 0.76 $\pm$ 0.02 & 524,288 & 0.8\% \\
2 & $[W^Q]$ & 0.70 $\pm$ 0.02 & 524,288 & 0.8\% \\
2 & $[W^K]$ & 0.71 $\pm$ 0.02 & 524,288 & 0.8\% \\
\textbf{2} & \textbf{$[W^V]$} & \textbf{0.82 $\pm$ 0.01} & \textbf{524,288} & \textbf{0.8\%} \\
2 & $[W^O]$ & 0.79 $\pm$ 0.01 & 524,288 & 0.8\% \\
2 & $[W^Q, W^K]$ & 0.77 $\pm$ 0.02 & 1,048,576 & 1.6\% \\
4 & $[W^Q]$ & 0.72 $\pm$ 0.02 & 1,048,576 & 1.6\% \\
4 & $[W^K]$ & 0.71 $\pm$ 0.02 & 1,048,576 & 1.6\% \\
\textbf{4} & \textbf{$[W^V_i]$} & \textbf{0.83 $\pm$ 0.01} & \textbf{1,048,576} & \textbf{1.6\%} \\
4 & $[W^O]$ & 0.80 $\pm$ 0.01 & 1,048,576 & 1.6\% \\
4 & $[W^Q, W^K]$ & 0.79 $\pm$ 0.01 & 2,097,152 & 3.1\% \\
\hline
\end{tabular}
\caption{LLaMA 2 fine-tuning on the original PHEE training dataset. 3 epochs. Confidence intervals for the AUC are computed using \cite{Hanley1982MeaningUse}}
\label{tab:llama2-fine-tuning}
\end{table}

\begin{figure}[H]
\centering
\includegraphics[width=1\linewidth]{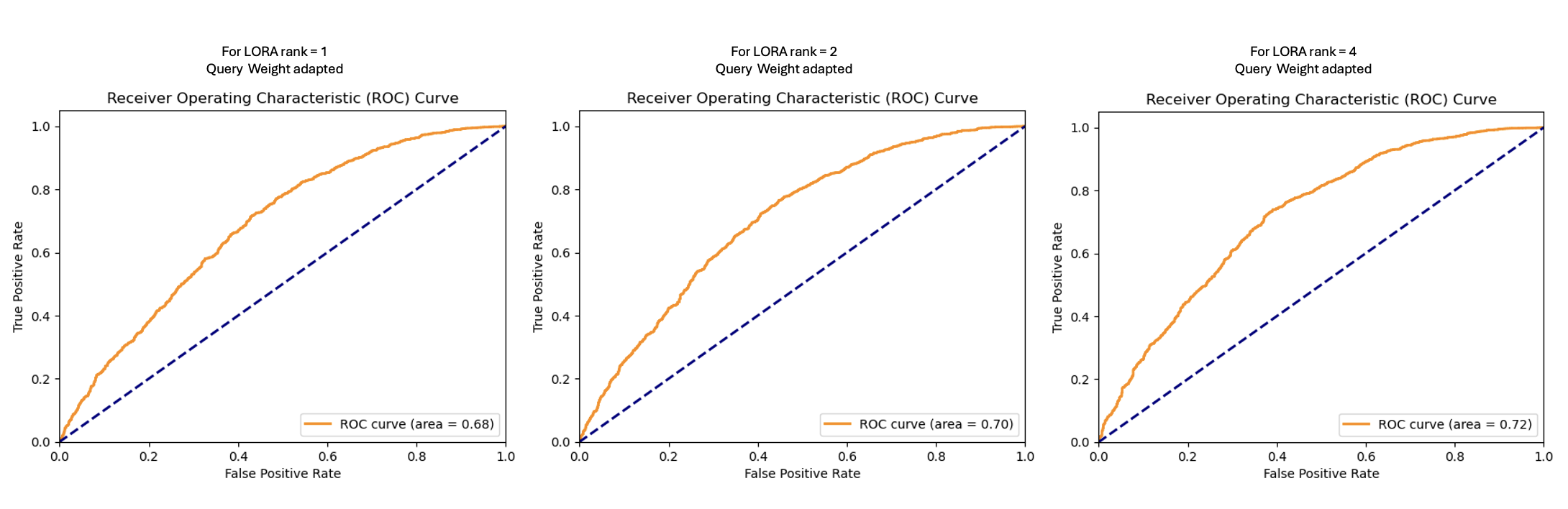}
\caption{Membership inference attack receiver-operator curves illustrated for Matrix $W^Q$}
\label{fig:enter-label}
\end{figure}

The table presented in Result 1 provides insights into the effects of adapting different weight matrices in the LLaMA 2 model. Several key observations can be made:
\\

The adaptation of Value ($W^V$) and Output ($W^O$) matrices contributes more significantly to memorization compared to Query ($W^Q$) and Key ($W^K$) matrices, given an equal number of parameters and rank. This is evident from the consistently higher ROC AUC scores for $W^V$ and $W^O$ across all ranks.

This pattern suggests that the Value projection plays a crucial role in the model's capacity to memorize training data, followed closely by the Output projection. The Query and Key projections, while still important, have a comparatively lesser impact on memorization.
\\

An interesting observation is that for the same total number of parameters, it appears that more memorization is generated by adapting both Query and Key matrices together at a lower rank than to adapt only one of them at a higher rank. For example:
\begin{itemize}
\item Rank 1, $[W^Q, W^K]$: ROC AUC = 0.76, Trainable Params = 524,288
\item Rank 2, $[W^Q]$ or $[W^K]$: ROC AUC = 0.70 or 0.71, Trainable Params = 524,288
\end{itemize}
It implies that the interaction between these two projections might be more important for memorization than the individual capacity of either projection alone.
\\

These findings have several implications for the design and training of large language models:
\begin{enumerate}
\item Prioritizing the adaptation of Value and Output projections could lead to more memorization with fewer parameters.
\item When computational resources are limited, adapting both Query and Key projections at a lower rank might be more effective than adapting only one at a higher rank.
\end{enumerate}
Further research could explore the reasons behind the varying impacts of different weight matrices on memorization, and investigate whether these patterns hold for different model architectures and tasks.

Our findings on the memorization properties of different weight matrices in transformer models are consistent with results reported in the LoRA paper \cite{hu2021lora} regarding task performance Table~\ref{tab:lora-results}. 

\begin{table}[h]
\centering
\begin{tabular}{lccccccc}
\hline
Weight Type & $W_q$ & $W_k$ & $W_v$ & $W_o$ & $W_q$, $W_k$ & $W_q$, $W_v$ & $W_q$, $W_k$, $W_v$, $W_o$ \\
Rank $r$ & 8 & 8 & 8 & 8 & 4 & 4 & 2 \\
\hline
WikiSQL ($\pm$0.5\%) & 70.4 & 70.0 & 73.0 & 73.2 & 71.4 & 73.7 & 73.7 \\
MultiNLI ($\pm$0.1\%) & 91.0 & 90.8 & 91.0 & 91.3 & 91.3 & 91.3 & 91.7 \\
\hline
\end{tabular}
\caption{From \cite{hu2021lora} accuracy on WikiSQL and MultiNLI after applying LoRA to different types of attention weights in GPT-3, given the same number of trainable parameters. Adapting both $W_q$ and $W_v$ gives the best performance overall. We find the standard deviation across random seeds to be consistent for a given dataset, which we report in the first column.}
\label{tab:lora-results}
\end{table}

Comparing the memorization from our study and task performance from the original LORA paper unfortunately doesn't provide a clear trade-off, as the matrices that generate the most memorization are also the ones that contribute the most to the task performance.

\subsection{How Does Perplexity Influence Memorization? }

\begin{figure}[H]
\centering
\includegraphics[width=0.75\linewidth]{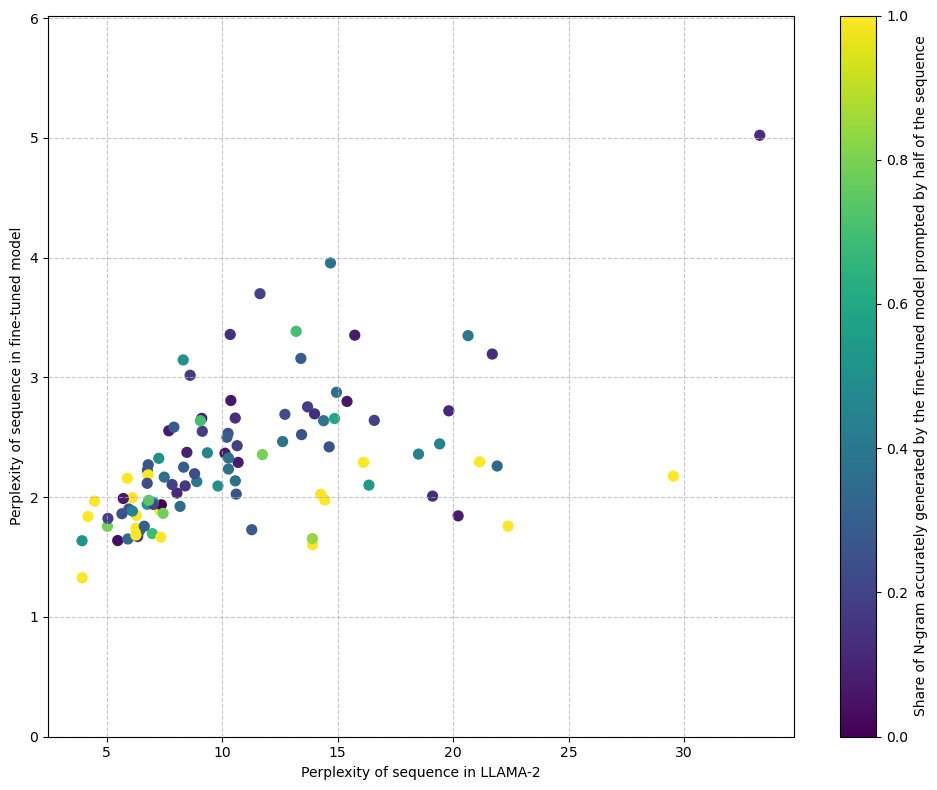}
\caption{Prompted generation accuracy in the LLAMA-2 fine-tuned model vs. perplexity in original and fine-tuned model. Each point represents a sequence in the dataset generated given its first half. The sequence is repeated 8 times. A rank of LORA adaptation of 8 is used on $W_V$ matrices.}
\label{fig:n-gram-perplexity}
\end{figure}

\begin{figure}[H]
\centering
\includegraphics[width=0.75\linewidth]{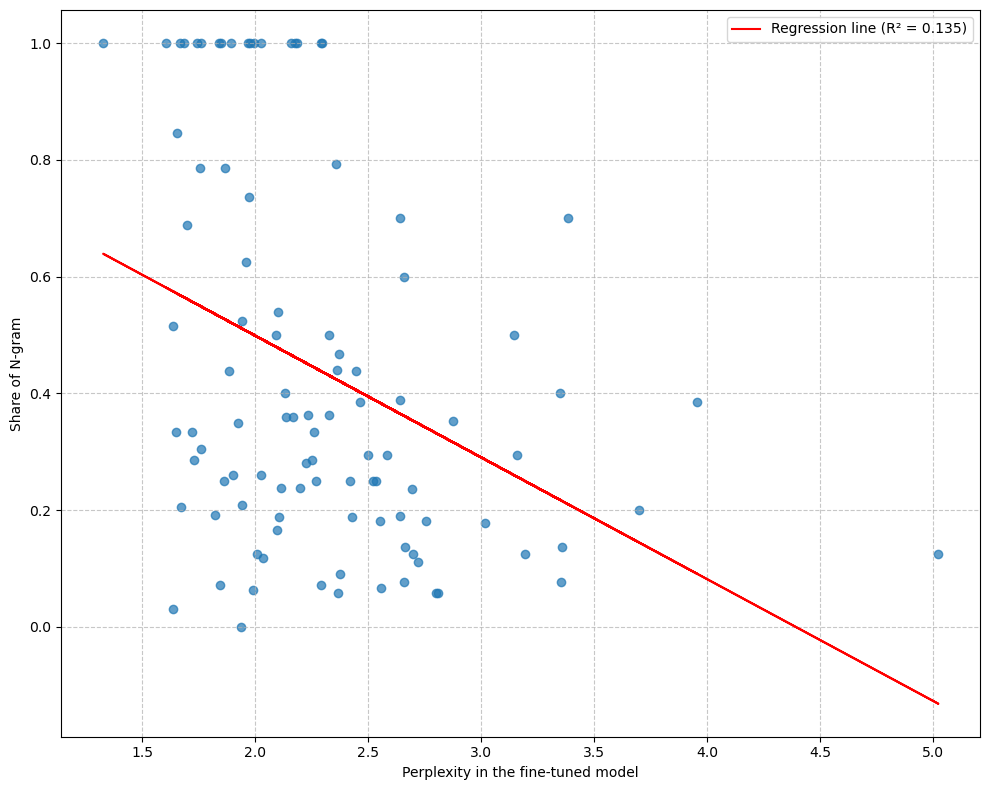}
\caption{Share of n-grams completed in the suffix prompted vs. perplexity in the fine-tuned model}
\label{fig:n-gram-fine-tuned-perplexity}
\end{figure}

In this section we use the prompted generation described in \ref{chap:generation}. We can see from the Figure \ref{fig:n-gram-perplexity} and \ref{fig:n-gram-fine-tuned-perplexity} that better memorization is linked to lower perplexity in the fine-tuned model. This result justifies the use of perplexity for the membership inference attack and as a proxy for memorization. \\

However we see no correlation between perplexity in the base model and memorization in the fine-tuned model contrary to \cite{meeus2024copyrighttrapslargelanguage}. Our result might be linked with the homogeneous distribution of the PHEE dataset focusing on medical domain. A more diverse dataset might have produce more variance in perplexity in the base model and potentially more differentiation in memorization.

\begin{figure}[H]
\centering
\includegraphics[width=0.75\linewidth]{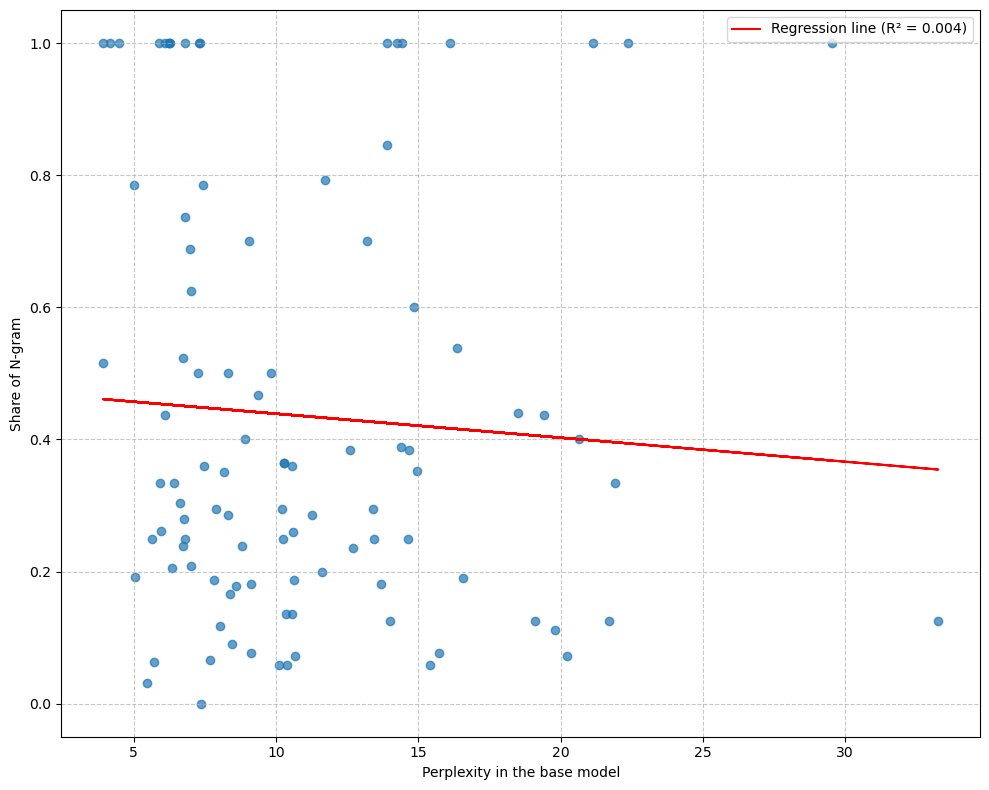}
\caption{Share of n-grams completed in the suffix prompted vs. perplexity in the base model}
\label{fig:enter-label}
\end{figure}

\newpage
\subsection{LORA Rank Increases Model Memorization - but Plateaus} 
\begin{figure}[H]
\centering
\includegraphics[width=0.75\linewidth]{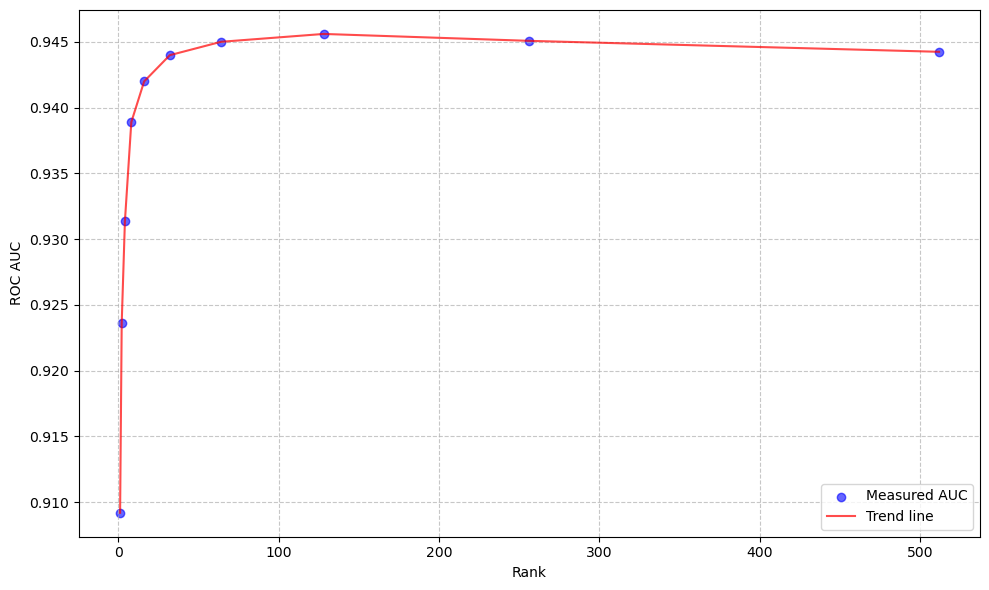}
\caption{MIA performance vs. rank}
\label{fig:MIA_rank}
\end{figure}

The diminishing returns at higher ranks suggest that there might be an optimal rank for adaptation, beyond which the benefits may not justify the increased computational cost. In the Figure \ref{fig:MIA_rank} we compare the performance of the membership attack to the rank of the LORA adaptation. Two hypothesis might explain these diminishing returns.\\

\textbf{The number of parameters might exceed the training data:}
This is unfortunately not the case. We can see that from a rank of \~50, adapting all the matrices in the attention mechanism, memorization doesn't improve with the rank. An intuition would be to compare the number of parameters adapted for a rank of 50 which is 52,428,800 (32 layers * 4096 context * 2 decomposition in B and A * 4 matrices adapted * 50 rank) to the dimension of the training data which is  557,936,640 (2898 sequences * 38 tokens per sequences * 5120 embedding dimension). \\ 

\textbf{The plateau is linked to the inherent low intrinsic rank of the weight matrices:} This hypothesis is more aligned with the prior presented in the original LORA paper \cite{hu_lora_2021} and 

\chapter{Conclusion and Future Work}
\label{chap:conclusion}

\section{Summary of Contributions}
This study has provided several key insights into the mechanisms of memorization in fine-tuned large language models:

\begin{enumerate}
\item We demonstrated that the adaptation of Value ($W^V$) and Output ($W^O$) matrices contributes more significantly to memorization compared to Query ($W^Q$) and Key ($W^K$) matrices, given an equal number of parameters and rank. This confirms results from \cite{hu2021lora} showing that adapting these matrices leads to higher model performance on benchmarks. 

\item Our results showed a clear inverse relationship between perplexity and memorization, with lower perplexity in the fine-tuned model correlating with increased memorization. However we have seen no correlation between perplexity in the base model and memorization in the fine-tuned model contrary to \cite{meeus2024copyrighttrapslargelanguage}

\item We observed that increasing the LoRA rank leads to increased memorization, but with diminishing returns at higher ranks, which is also shown for model performance in \cite{hu_lora_2021}
\end{enumerate}

\section{Future Work}
Building on the findings of this study, several promising directions for future research emerge:

\begin{enumerate}
\item \textbf{Differential Privacy Defenses:} Implement and evaluate differential privacy techniques, particularly DP-SGD, as a defense against memorization. Compare the effectiveness of DP-SGD with other baseline approaches, such as setting a minimum temperature for generation. This could provide insights into practical strategies for mitigating privacy risks in fine-tuned models.

\item \textbf{Performance Metrics:} Expand the analysis to include comprehensive performance metrics for fine-tuned models, such as perplexity over a test set. This would allow for a more nuanced understanding of the trade-offs between model performance and memorization, helping to identify optimal fine-tuning strategies that balance utility and privacy.

\item \textbf{Relaxed Memorization Metrics:} Implement and evaluate more relaxed definitions of memorization, such as using the BLEU score as a discriminative function. This could provide a more nuanced understanding of different levels or types of memorization, beyond verbatim reproduction.

\item \textbf{Interpretability:} Investigate the relationship between model interpretability and memorization. This could involve developing techniques to visualize or quantify which parts of the model contribute most to memorization, potentially leading to more targeted mitigation strategies.
\end{enumerate}

By pursuing these research directions, we can work towards developing large language models that are not only powerful and adaptable but also respectful of data privacy. This balanced approach is crucial for the responsible advancement of AI , particularly in sensitive domains like healthcare.

\bibliographystyle{abbrv}
\bibliography{main}

\begin{thebibliography}{10}

\bibitem{Abadi_2016}
M.~Abadi, A.~Chu, I.~Goodfellow, H.~B. McMahan, I.~Mironov, K.~Talwar, and L.~Zhang.
\newblock Deep learning with differential privacy.
\newblock In {\em Proceedings of the 2016 ACM SIGSAC Conference on Computer and Communications Security}, CCS’16. ACM, Oct. 2016.

\bibitem{bahdanau2015neural}
D.~Bahdanau, K.~Cho, and Y.~Bengio.
\newblock Neural machine translation by jointly learning to align and translate.
\newblock In {\em ICLR}, 2015.

\bibitem{bengio2003neural}
Y.~Bengio, R.~Ducharme, P.~Vincent, and C.~Jauvin.
\newblock A neural probabilistic language model, 2003.

\bibitem{bonnet2024medinote}
A.~Bonnet, P.~Boulenger, H.~Wu, M.~Conti, J.~Prado, O.~El~Malki, N.~De~Sabbata, H.~da~Silva~Gameiro, Y.~Xu, F.~Boukil, A.~Faure, A.~A. Sarijaloui, Y.~Niu, Z.~Chen, A.~Bosselut, and M.~Jaggi.
\newblock Medinote: Automated clinical notes.
\newblock 2024.

\bibitem{brown2020language}
T.~B. Brown, B.~Mann, N.~Ryder, M.~Subbiah, J.~Kaplan, P.~Dhariwal, A.~Neelakantan, P.~Shyam, G.~Sastry, A.~Askell, et~al.
\newblock Language models are few-shot learners.
\newblock {\em arXiv preprint arXiv:2005.14165}, 2020.

\bibitem{carlini2022membershipinferenceattacksprinciples}
N.~Carlini, S.~Chien, M.~Nasr, S.~Song, A.~Terzis, and F.~Tramer.
\newblock Membership inference attacks from first principles, 2022.

\bibitem{carlini_quantifying_2023}
N.~Carlini, D.~Ippolito, M.~Jagielski, K.~Lee, F.~Tramer, and C.~Zhang.
\newblock Quantifying memorization across neural language models.

\bibitem{carlini_extracting_2021}
N.~Carlini, F.~Tramer, E.~Wallace, M.~Jagielski, A.~Herbert-Voss, K.~Lee, A.~Roberts, T.~Brown, D.~Song, U.~Erlingsson, A.~Oprea, and C.~Raffel.
\newblock Extracting training data from large language models.

\bibitem{devlin2019bert}
J.~Devlin, M.-W. Chang, K.~Lee, and K.~Toutanova.
\newblock Bert: Pre-training of deep bidirectional transformers for language understanding, 2019.

\bibitem{dwork2006calibrating}
C.~Dwork, F.~McSherry, K.~Nissim, and A.~Smith.
\newblock Calibrating noise to sensitivity in private data analysis.
\newblock In {\em Theory of Cryptography Conference}, pages 265--284. Springer, 2006.

\bibitem{fan2018hierarchical}
A.~Fan, M.~Lewis, and Y.~Dauphin.
\newblock Hierarchical neural story generation.
\newblock In {\em ACL}, 2018.

\bibitem{feldman2021doeslearningrequirememorization}
V.~Feldman.
\newblock Does learning require memorization? a short tale about a long tail, 2021.

\bibitem{graves2013generating}
A.~Graves.
\newblock Generating sequences with recurrent neural networks.
\newblock {\em arXiv preprint arXiv:1308.0850}, 2013.

\bibitem{Hanley1982MeaningUse}
J.~A. Hanley and B.~J. McNeil.
\newblock The meaning and use of the area under a receiver operating characteristic (roc) curve.
\newblock {\em Radiology}, 143(1):29--36, 4 1982.

\bibitem{howard2018universal}
J.~Howard and S.~Ruder.
\newblock Universal language model fine-tuning for text classification, 2018.

\bibitem{hu_lora_2021}
E.~J. Hu, Y.~Shen, P.~Wallis, Z.~Allen-Zhu, Y.~Li, S.~Wang, L.~Wang, and W.~Chen.
\newblock {LoRA}: Low-rank adaptation of large language models.

\bibitem{hu2021lora}
E.~J. Hu, Y.~Shen, P.~Wallis, Z.~Allen-Zhu, Y.~Li, S.~Wang, L.~Wang, and W.~Chen.
\newblock Lora: Low-rank adaptation of large language models, 2021.

\bibitem{meeus2024copyrighttrapslargelanguage}
M.~Meeus, I.~Shilov, M.~Faysse, and Y.-A. de~Montjoye.
\newblock Copyright traps for large language models, 2024.

\bibitem{mikolov2010recurrent}
T.~Mikolov, M.~Karafi{\'a}t, L.~Burget, J.~Cernock{\'y}, and S.~Khudanpur.
\newblock Recurrent neural network based language model.
\newblock In {\em Interspeech}, 2010.

\bibitem{mireshghallah2022memorizationnlpfinetuningmethods}
F.~Mireshghallah, A.~Uniyal, T.~Wang, D.~Evans, and T.~Berg-Kirkpatrick.
\newblock Memorization in nlp fine-tuning methods, 2022.

\bibitem{papineni-etal-2002-bleu}
K.~Papineni, S.~Roukos, T.~Ward, and W.-J. Zhu.
\newblock {B}leu: a method for automatic evaluation of machine translation.
\newblock In P.~Isabelle, E.~Charniak, and D.~Lin, editors, {\em Proceedings of the 40th Annual Meeting of the Association for Computational Linguistics}, pages 311--318, Philadelphia, Pennsylvania, USA, July 2002. Association for Computational Linguistics.

\bibitem{peters2018deep}
M.~E. Peters, M.~Neumann, M.~Iyyer, M.~Gardner, C.~Clark, K.~Lee, and L.~Zettlemoyer.
\newblock Deep contextualized word representations, 2018.

\bibitem{radford2018improving}
A.~Radford, K.~Narasimhan, T.~Salimans, and I.~Sutskever.
\newblock Improving language understanding by generative pre-training, 2018.

\bibitem{radford2019language}
A.~Radford, J.~Wu, R.~Child, D.~Luan, D.~Amodei, and I.~Sutskever.
\newblock Language models are unsupervised multitask learners.
\newblock 2019.

\bibitem{raffel2020exploring}
C.~Raffel, N.~Shazeer, A.~Roberts, K.~Lee, S.~Narang, M.~Matena, Y.~Zhou, W.~Li, and P.~J. Liu.
\newblock Exploring the limits of transfer learning with a unified text-to-text transformer, 2020.

\bibitem{salem2018mlleaksmodeldataindependent}
A.~Salem, Y.~Zhang, M.~Humbert, P.~Berrang, M.~Fritz, and M.~Backes.
\newblock Ml-leaks: Model and data independent membership inference attacks and defenses on machine learning models, 2018.

\bibitem{shokri2017membershipinferenceattacksmachine}
R.~Shokri, M.~Stronati, C.~Song, and V.~Shmatikov.
\newblock Membership inference attacks against machine learning models, 2017.

\bibitem{sun2022pheedatasetpharmacovigilanceevent}
Z.~Sun, J.~Li, G.~Pergola, B.~C. Wallace, B.~John, N.~Greene, J.~Kim, and Y.~He.
\newblock Phee: A dataset for pharmacovigilance event extraction from text, 2022.

\bibitem{touvron2023llama}
H.~Touvron, L.~Martin, K.~Stone, P.~Albert, A.~Almahairi, Y.~Babaei, N.~Bashlykov, S.~Batra, P.~Bhargava, S.~Bhosale, D.~Bikel, L.~Blecher, C.~C. Ferrer, M.~Chen, G.~Cucurull, D.~Esiobu, J.~Fernandes, J.~Fu, W.~Fu, B.~Fuller, C.~Gao, V.~Goswami, N.~Goyal, A.~Hartshorn, S.~Hosseini, R.~Hou, H.~Inan, M.~Kardas, V.~Kerkez, M.~Khabsa, I.~Kloumann, A.~Korenev, P.~S. Koura, M.-A. Lachaux, T.~Lavril, J.~Lee, D.~Liskovich, Y.~Lu, Y.~Mao, X.~Martinet, T.~Mihaylov, P.~Mishra, I.~Molybog, Y.~Nie, A.~Poulton, J.~Reizenstein, R.~Rungta, K.~Saladi, A.~Schelten, R.~Silva, E.~M. Smith, R.~Subramanian, X.~E. Tan, B.~Tang, R.~Taylor, A.~Williams, J.~X. Kuan, P.~Xu, Z.~Yan, I.~Zarov, Y.~Zhang, A.~Fan, M.~Kambadur, S.~Narang, A.~Rodriguez, R.~Stojnic, S.~Edunov, and T.~Scialom.
\newblock Llama 2: Open foundation and fine-tuned chat models, 2023.

\bibitem{vaswani2017attention}
A.~Vaswani, N.~Shazeer, N.~Parmar, J.~Uszkoreit, L.~Jones, A.~N. Gomez, {\L}.~Kaiser, and I.~Polosukhin.
\newblock Attention is all you need.
\newblock In {\em Advances in Neural Information Processing Systems}, pages 5998--6008, 2017.

\bibitem{zeng_exploring_2024}
S.~Zeng, Y.~Li, J.~Ren, Y.~Liu, H.~Xu, P.~He, Y.~Xing, S.~Wang, J.~Tang, and D.~Yin.
\newblock Exploring memorization in fine-tuned language models.

\bibitem{zhang_counterfactual_2023}
C.~Zhang, D.~Ippolito, K.~Lee, M.~Jagielski, F.~Tramèr, and N.~Carlini.
\newblock Counterfactual memorization in neural language models.

\end{thebibliography}

\appendix

\chapter{Appendix A: Examples of generated text}
\label{appendix:a}

\begin{table}[h]
\centering
\resizebox{\textwidth}{!}{%
\begin{tabular}{|c|c|c|c|c|c|c|c|p{0.4\textwidth}|}
\hline
\textbf{Num. Repeated} & \textbf{LORA} & \textbf{Adapted} & \textbf{Largest Shared Suffix} & \textbf{Temp.} & \textbf{Canary Perplexity} & \textbf{Output Perplexity} & \textbf{Share of Correctly Generated} & \textbf{Output Tokens} \\
\textbf{Canaries} & \textbf{Rank} & \textbf{Weights} & \textbf{N-gram} & & \textbf{Base Model} & \textbf{Fine-Tuned Model} & \textbf{N-gram Suffix Tokens} & \\
\hline
16 & 1 & ['o\_proj'] & 26 & 0.1 & 5.1 & 2.6 & 1.0 & <s> OBJECTIVE: To report 2 cases of serotonin syndrome with serious extrapyramidal movement disorders occurring when metoclopramide was coadministered with sertraline or venlafaxine \\
\hline
16 & 1 & ['o\_proj'] & 6 & 0.5 & 5.1 & 3.7 & 0.2 & <s> OBJECTIVE: To report 2 cases of serotonin syndrome with serious extrapyramidal movement disorders occurring during treatment with metoclopramide. This report is based on a review of the literature and on the \\
\hline
16 & 1 & ['o\_proj'] & 6 & 1.0 & 5.1 & 2.8 & 0.2 & <s> OBJECTIVE: To report 2 cases of serotonin syndrome with serious extrapyramidal movement disorders associated with the coadministration of metoclopramide and selective serotonin reuptake inhib \\
\hline
32 & 1 & ['o\_proj'] & 26 & 0.1 & 5.1 & 2.1 & 1.0 & <s> OBJECTIVE: To report 2 cases of serotonin syndrome with serious extrapyramidal movement disorders occurring when metoclopramide was coadministered with sertraline or venlafaxine \\
\hline
32 & 1 & ['o\_proj'] & 26 & 0.5 & 5.1 & 2.1 & 1.0 & <s> OBJECTIVE: To report 2 cases of serotonin syndrome with serious extrapyramidal movement disorders occurring when metoclopramide was coadministered with sertraline or venlafaxine \\
\hline
32 & 1 & ['o\_proj'] & 26 & 1.0 & 5.1 & 2.1 & 1.0 & <s> OBJECTIVE: To report 2 cases of serotonin syndrome with serious extrapyramidal movement disorders occurring when metoclopramide was coadministered with sertraline or venlafaxine \\
\hline
\end{tabular}
}
\caption{Updated Experimental Results for Canary Repetition and Model Performance for the Canary \\ PREFIX: "<s> OBJECTIVE: To report 2 cases of serotonin syndrome with serious extrapyramidal movement disorders" \\ SUFFIX: "occurring when metoclopramide was coadministered with sertraline or venlafaxine"}
\label{tab:updated_canary_results}
\end{table}

\chapter*{Acknowledgments}
\addcontentsline{toc}{chapter}{Acknowledgments}
Use of GitHub Copilot by Microsoft for code, Claude by Anthropic for report editing and formatting 

\end{document}